\documentclass[letterpaper, 10 pt, conference]{ieeeconf}  

\IEEEoverridecommandlockouts % This command is only needed if 
\usepackage{hyperref}
\usepackage[vertfit]{breakurl} 
\usepackage{tabularx}
\usepackage{flushend}

\title{\LARGE \bf Expanding the Set of Pragmatic Considerations in Conversational AI}
\author{S.M. Seals$^{1,2,3}$ and Valerie L. Shalin$^{3,4}$% <-this % stops a space
\thanks{$^{1}$Air Force Research Laboratory, $^{2}$Oak Ridge Institute for Science and Education, $^{3}$Wright State University, $^{4}$Artificial Intelligence Institute- University of South Carolina. The views expressed are those of the authors and do not necessarily reflect the official policy or position of the Department of the Air Force, the Department of Defense, or the U.S. government. Approved for public release, case number: AFRL20225175.}
\thanks{S.M. Seals: {\tt\small s.m.seals @ outlook.com}}}

\begin{document}

\maketitle
\thispagestyle{empty}
\pagestyle{empty}

\begin{abstract}

Despite considerable performance improvements, current conversational AI systems often fail to meet user expectations. We discuss several pragmatic limitations of current conversational AI systems. We illustrate pragmatic limitations with examples that are syntactically appropriate, but have clear pragmatic deficiencies. We label our complaints as "Turing Test Triggers" (TTTs) as they indicate where current conversational AI systems fall short compared to human behavior. We develop a taxonomy of pragmatic considerations intended to identify what pragmatic competencies a conversational AI system requires and discuss implications for the design and evaluation of conversational AI systems.
\end{abstract}

\section{Introduction}

Advances in deep learning and large language models have enabled the development of high performing NLP and conversational applications \cite{radford_language_2019, brown_language_2020, devlin_bert_2019, zhangOPTOpenPretrained2022, ouyangTrainingLanguageModels2022}. This work has %enabled the development of 
yielded conversational AI applications that appear to reflect the characteristics of human dialogue and follow user instructions \cite{kim_will_2020, majumder_mime_2020, wu_proactive_2019, ouyangTrainingLanguageModels2022}. Performance improvements have prompted new empirical work on evaluation (i.e., \cite{bhandari_re-evaluating_2020, meister_language_2021, wang_perplexity_2022}). 
In that spirit, we illustrate several current challenges for conversational AI systems. We illustrate these limitations with examples from conversational AI systems in the literature \cite{gratch_distress_2014, Manas2021} and author interactions with currently fielded conversational AI systems \footnote{Chatbots: \href{https://openai.com/}{OpenAI's chatGPT}, \href{https://www.amtrak.com/home}{Amtrak's Julie}, \href{https://woebothealth.com/try-woebot/}{WoebotHealth's Woebot}} and voice assistants\footnote{Voice assistants: \href{https://www.apple.com}{Apple's Siri}, \href{https://alexa.amazon.com}{Amazon's Alexa}}. These examples are syntactically appropriate, but have clear pragmatic deficiencies compared to human behavior. This discrepancy triggers the Turing Test criterion- competent human speakers and users would not produce such constructions. We draw on traditional (i.e., travel, personal assistants) and more recent (i.e., LLM chatbot interfaces, mental health applications). Chatbots clearly emphasize some of our concerns and, particularly for medical applications, require highly refined performance. 

We structure documented general user frustrations with conversational AI systems that highlight complaints about conversational skills separately from other usability concerns \cite{brandtzaeg_why_2017, folstad_chatbots_2019,liao_what_2016, luger_like_2016, porcheron_voice_2018, zamora_im_2017}. In so doing, we synergize applied and basic research endeavors that address language in use. Users, particularly in consequential task domains, are less tolerant of limitations than researchers. 

We frame pragmatic limitations (and resulting user frustrations) of current conversational AI systems using socially-inspired pragmatic theory of \emph{relevance}  \cite{Wilson2013}. We articulate two sub-themes for understanding and addressing these limitations: preserving local meaning and incorporating context. The resulting taxonomy informs pragmatic criteria for designing and evaluating conversational AI systems, and integrates insights from social and behavioral sciences with computational science.  

\section{User Relevance}
Conversations preserve \emph{relevance}. When people engage in conversation, they expect their partners will make relevant contributions that are consistent with the accepted purpose of the conversation. Users have similar expectations for interactions with conversational AI applications (i.e., \cite{zamora_im_2017}). \cite{Grice1975} initially proposed that the expectation of relevance is due to a \emph{cooperative principle} and that the expectation of related utterances is due to a maxim of \emph{relation}. 

\cite{sperber_relevance_1986} revised this explanation and proposed the search for relevance as a basic feature of human cognition \cite{Wilson2013}. Input is relevant if the processing it generates a worthwhile change in a recipient's representation of the world \cite{Wilson2013}. Relevance depends on context. Information must be worth the recipient's processing effort and be the most relevant information available consistent with their goals \cite{Wilson2013}. 
Content from a conversational AI system that is incorrect, difficult to understand, or missing important information reduces relevance. Users must expend additional comprehension effort or search elsewhere. Consider this example (originally proposed by \cite{davis_experiments_2022}, repeated on chatGPT May 24 2023 Version):

\begin{quote}
Prompt: \emph{You need flour to bake bread. You have a sack of flour in the garage. When you get there, you find that the flour is at the bottom of the sack, but that somebody emptied a couple of boxes of dishwashing detergent on top of it. So you have to...}

(TTT) Response: \emph{carefully remove the boxes of dishwashing detergent from on top of the sack of flour, making sure not to spill or mix any of the detergent with the flour. You can gently lift the boxes, taking caution to avoid any potential spills or mess. Once the boxes are safely removed, you can access the flour at the bottom of the sack and proceed to use it for baking your bread, ensuring it remains uncontaminated by the dishwashing detergent.} 
\end{quote}

The response acknowledges that you should not make bread with flour contaminated by dishwashing detergent. However, the response fails to account for the fact that the dishwashing detergent has been \emph{emptied into} the flour, violating the expectation of providing all relevant implications, and resulting in a dangerous recommendation.

The example above, and related empirical work in HCI, illustrates two dimensions of relevance; local meaning and contextual awareness. Pragmatically sensitive responses require integration with these dimensions. Next, we examine each dimension with particular focus on subcategories of contextual awareness.

\section{Meaning and Inference}

To generate relevant content, conversational AI systems must respond to all aspects of a user's meaning. A response that addresses one part of a user’s intent may omit other related information. This requires recovering the complex relationships within an utterance \cite{forster_visual_1970, graesser_structural_1980}. Systems that are unable to account for these relationships compromise coherence and require additional user effort that impairs comprehension \cite{beck_revising_1991, mcnamara_are_1996}. Consider this voice assistant example:

\begin{quote}
    User: \emph{I want to go to Cleveland, is there any construction that would slow down my trip?}
    
    (TTT) Assistant: \emph{Getting directions to Cleveland [does not provide traffic information].}
    
    Pragmatically-appropriate assistant: \emph{Traveling to Cleveland will take 3 hours. There are currently no traffic delays. [Pulls up directions]}
\end{quote}

The assistant answers the first aspect of the request, directions to a given city. The second aspect, travel delays, is ignored. Pragmatically appropriate responses require a representation of both the requested information and inference. The user is concerned about travel time- construction is merely one example. %Such examples resolve with better representations of the  individual utterance.

\section{Context}
A broader category of pragmatic failures concerns the failure to address other context \cite{talmy_cognitive_2000}. The first is conversational context, which addresses relationships between utterances and the overall conversation. The second concerns factors external to the conversation. 

Psychologists invoke the construct of memory to explain context-related processing. Semantic memory contains general knowledge while episodic memory addresses specific events and associated details \cite{nyberg_general_1996, tulving_episodic_1972}. Both memory sources influence conversational and external context. 

\subsection{Conversational Context}
Generating relevant content requires systems that can incorporate the meaning of an utterance with respect to the entire conversation. Specific topics, words, and phrases create a conversational context that influences later comprehension and production. People represent this context and recursively combine and update it throughout the conversation \cite{clark_common_1983, Isaacs1987, Lockridge2002, VanDijk1983}. 

Results from HCI research demonstrate that users want systems that maintain conversational context \cite{liao_what_2016, luger_like_2016, zamora_im_2017} and consider conversational abilities more important than other features \cite{folstad_chatbots_2019}. The ideal collaborative system demonstrates awareness of the user's status and intentions in responding \cite{liao_what_2016, folstad_chatbots_2019}. Pragmatic failures arise when conversational AI systems fail to incorporate these aspects of human conceptual representations. 

\subsubsection{Awareness of Dialogue Topics}

Speakers in conversation create, modify, and update expectations about what all participants know \cite{clark1996using, pickering_toward_2004}. These representations include information from personal experience, community membership, and conversational context. Conversational AI applications either do not create these representations (i.e., \cite{rollo_carpenterexistor_cleverbot_2022}) or do so in a limited fashion \cite{ouyangTrainingLanguageModels2022}. Failure to account for this history generates pragmatically inappropriate responses. Consider this patient interview system: 

\begin{quote}
    (TTT) Interviewer: \emph{Do you \underline{still} go to therapy now?}
    
    Patient: \emph{I don't, I \underline{never} went to therapy.}
\end{quote}

The word \emph{still} presupposes previous therapy, without having previously established that fact. An appropriate exchange would first establish whether the patient has attended therapy and, conditional on an affirmative response, ask if the patient still attends therapy. 

\subsubsection{Propose Relevant Next Turns}
Similarly, systems must account for the prior conversation to generate responses. This challenge is particularly apparent for extended interactions. The example below is from a mental health chatbot. The chatbot poses questions intended to help the user process a thought.  

\begin{quote}
    Bot: \emph{Does this thought assume others are thinking negatively about you?}
    
    User: \emph{No}
    
    Bot: \emph{Does your thought place too much blame on others?}
    
    User: \emph{No it doesn't}
    
    Bot: \emph{Does it place too much blame on you?}
    
    User: \emph{Nope}
    
    (TTT) Bot: \emph{Can you rewrite the thought without any distortions?}
    
    User: \emph{It doesn't have any}

    (TTT) Bot: \emph{How did you do?}
\end{quote}
 
The last two questions ignore the history of the conversation. An appropriate response could introduce a new strategy or propose an alternative action.

\subsubsection{Correction of Prior Content}

Dialogue is a collaborative process that requires updating and correcting information \cite{clark_collaborating_1987, brennan_conceptual_1996, Schegloff1977}. Despite their pervasiveness, conversational AI systems struggle with these phenomena:

\begin{quote}
    User: \emph{I want to book a ticket to Newark, sorry New York}
    
    (TTT) Bot: \emph{There are multiple destinations, please select from the list below. [list includes Newark, NJ, Newark DE, and Newark NJ.]}
\end{quote}

This response does not address the user’s self-correction. Appropriate responses would suggest New York, as intended, or request clarification. Previous work has examined clarification generation (i.e., \cite{benotti_grounding_2021, kim_deciding_2021, majumder_ask_2021}. The challenge is identifying when clarification is required. 

\subsection{External Context} 
Failure to account for external information can generate pragmatic failures. Speakers in conversation expect that their partners possess certain background knowledge and awareness of the physical environment \cite{Clark1981}. People exploit context to provide appropriate detail and tailor their messages \cite{clark_common_1983, Isaacs1987, van_der_henst_testing_2004}. Users want conversational AI systems to account for external context and find it frustrating when systems cannot \cite{liao_what_2016, luger_like_2016}. Where prior work has incorporated context, the focus has been on the conversation (i.e., \cite{ouyangTrainingLanguageModels2022, shuster_blenderbot_2022}). We suggest this approach is insufficient for truly collaborative conversational AI systems. While certainly a challenging goal, systems that incorporate external context will be easier to use and more consistent with users’ expectations.

\subsubsection{Episodic Features}
Speakers regularly use words and expressions that require context to be understood (i.e., \emph{today}). These features are easy to articulate and interpret in conversation- the current context or non-verbal cues suggest the intended meaning. This information is traditionally represented in different groups \cite{barwise_situations_1983}- facts a third party could observe about the conversation, information all speakers know, and facts that could verify or falsify aspects the conversation. Our taxonomy incorporates these contextual features that are often omitted from conversational AI systems.

Imprecise language creates an interpretation challenge for conversational AI systems. Systems often rely only on conversational content and neglect relevant information from the environment and external knowledge sources. Conversational AI systems struggle with conceptual and commonsense knowledge \cite{davis_commonsense_2015, speer_conceptnet_2017}, domain specific content \cite{gaur__2018}, attribute information \cite{zhang_collaborative_2016}, and prior information about the user. Systems that account for certain aspects require users to explicitly state what information should be considered \cite{ouyangTrainingLanguageModels2022}. For example, when interacting with chatGPT or GPT-4, users must explicitly state their preferred response style.

A lack of external context awareness generates several problems for conversational AI. Consider the following voice assistant example: 
\begin{quote}
    User: \emph{Is there a heat warning today?}
    
    (TTT) Assistant: \emph{I found this on the web} [Provides news article about heat wave in the UK when the user is in the US.]
    
    Pragmatically-appropriate assistant: \emph{Yes, there is a heat warning effect in [area] until [time].}
\end{quote}

As is typical for conversation, the user does not specify their location. The voice assistant fails to incorporate this information and generates a response inconsistent with the user's intent. The second response is more appropriate- it provides relevant and tailored information. Indeed, users have expectations for information that conversational AI systems should have and want systems to use that information \cite{luger_like_2016}.

\subsubsection{Conceptual Knowledge}
Interactions between a lack of external awareness and relevant semantic knowledge can generate pragmatic failures:

\begin{quote}
    User: \emph{Do I need gloves today?}
    
    (TTT) Assistant: \emph{Not much sun in the forecast today.}
    
    Pragmatically-appropriate assistant: \emph{The temperature is X degrees today.}
\end{quote}

An appropriate response requires knowledge of the current situation- the user’s location, current day, and weather forecast \cite{Levinson}. Moreover, the question requires conceptual knowledge. An appropriate answer accounts for the purpose of gloves (i.e., commonsense reasoning \cite{davis_commonsense_2015}). The response demonstrates awareness of the current situation, but omits required conceptual knowledge. An appropriate answer would include the forecasted low temperature. Users want systems that can account for these intentions and respond accordingly \cite{liao_what_2016}.

\subsubsection{Default Reasoning}
Incomplete information often requires the ability to draw conclusions based on general principles or identify when new information invalidates old conclusions \cite{brewka_introduction_2012}. Conversation regularly invokes these abilities \cite{van_der_henst_truthfulness_2002}, yet these situations pose problems for conversational AI applications. Consider a modification of a prior example: 

\begin{quote}
    User: \emph{I want to go to Cleveland, are there any traffic delays?}
    
    (TTT) Assistant: \emph{Getting directions to Cleveland [does not provide information about delays].}
    
    Pragmatically-appropriate assistant: \emph{Traveling to Cleveland will take 3 hours. There are no current delays. [Pulls up directions]}
\end{quote}

A pragmatically appropriate response would acknowledge all likely sources of traffic delays. While construction is the most prototypical, an appropriate response would account for other potential delays (i.e., a high probability snowstorm). Similarly, an appropriate response accounts for the probability that a situation will become relevant. Warnings about minor slowdowns several hours ahead would be pragmatically inappropriate. 

\begin{table}[h]
\caption{\label{pragmatics-guide} Considering pragmatic requirements for conversational AI systems.}

\begin{tabularx}{0.99\columnwidth}{|X|}

\hline
\textbf{Local Propositional Content:} Can the system address multiple propositions? Can the system incorporate previous content to create responses? Does this ability have a time frame and is the time frame appropriate? Can the system handle common sources of semantic imprecision? Are there methods for addressing unclear content? How are they formulated?
\\
\textbf{Distal Propositional Content:} Does this application require representing information across sentences or turns? What time frame should apply? Can the system identify inconsistent details? Can prior information be corrected? What correction strategies exist? \\
\textbf{Access to the External Environment:} Does the system represent details of the current situation? Can this information be meaningfully integrated with dialogue? What information about the environment would be helpful? Is other contextual information required? \\
\textbf{Access to External Knowledge:} What knowledge, general or domain specific, does the system require? How is it integrated? Can the system draw general conclusions or identify inconsistent details? \\
\hline
\end{tabularx}

\end{table}

\subsubsection{Inconsistent Details}

Similarly, conversations often require reasoning with inconsistent details \cite{brewka_introduction_2012}. Inconsistent details require %reasoners to 
the identification of inconsistent information and determination of what to disregard.

Humans resolve inconsistent details effectively \cite{johnson-laird_reasoning_2004}, but they create challenges for conversational AI systems. Systems that lack these abilities create pragmatic errors:

\begin{quote}
    User: \emph{Remind me on Friday August 4th at 5:00 to order groceries. [Friday is August 5th, not August 4th]}
    
    (TTT) Assistant: \emph{Done [creates reminder for Thursday August 4th at 5:00]}
    
    Pragmatically-appropriate assistant: \emph{Did you mean Thursday August 4th or Friday August 5th?}
\end{quote}

This requires detecting the inconsistency between the \emph{Friday} and \emph{the 4th} and resolve what the user intended. An appropriate response requires the ability to request clarification. Failure to detect and resolve inconsistent information results in conversational breakdown \cite{ashktorab_resilient_2019}. Inconsistent information is compounded in situations where dialogue accompanies real world activity (such as in meetings). 

Previous work has proposed methods for generating clarification requests when conversational AI systems are unsure of a user’s intent (i.e., \cite{benotti_recipe_2021, kim_deciding_2021, majumder_ask_2021}). Given that discrepancies have been adequately identified, similar methods could be used to resolve inconsistencies created by inconsistent details.

\subsubsection{Expert Knowledge}

Domain specific applications are not immune from external context pragmatic failures.%environments. 
These applications require conversational AI systems with appropriate background knowledge that generate appropriate responses for the intended audience \cite{Bell1984, clark1996using}. For example, defining new anatomy terms is appropriate for automated tutoring systems, but unnecessary in a personal assistant for physicians. Similarly, conversational AI systems need an awareness of domain content when intended for domain-specific applications (i.e., \cite{ferreira_profanity_2020}).

\section{Discussion}
We have shown that several limitations of current conversational AI systems are symptomatic of a more general problem: a lack of attention to pragmatics. We propose pragmatic failures are captured by \emph{relevance theory} \cite{Wilson2013}, and suggest two key limitations for conversational AI systems: preserving meaning and awareness of external context. We compile our concerns into a guide (Table ~\ref{pragmatics-guide}) designed to assess pragmatic requirements for a given application and the sufficiency of proposed strategies. 

Some of the ideas here have been examined in cooperative responding (i.e., \cite{cheikes_elements_1989}). However, these issues are not resolved with respect to modern deep learning based conversational AI systems. Previous work examining pragmatics has primarily investigated specific pragmatic features independently for specific applications (i.e., \cite{pandia_pragmatic_2021, ettinger_what_2020, gubelmann_context_2022, wang_calibrate_2021, nie_pragmatic_2020, schuz_decoupling_2021, zhang_improving_2022, bao_learning_2022, kim_will_2020, kim_perspective-taking_2021, Nath2020, wu_pragmatically_2021}. Treating pragmatics as a decentralized process ignores the interdependent nature of many pragmatic limitations. While resolving one of these issues may improve performance, truly context sensitive systems require the ability to address multiple issues. Some of the limitations we discuss are more glaringly obvious than others (i.e., systems that fail to recover local propositional content). However, all contribute to the design of truly cooperative and context-sensitive conversational AI systems. We suggest that the greatest challenge to creating pragmatically appropriate conversational AI systems is designing centralized systems that address multiple pragmatic limitations.

Recent research is addressing some of the issues we discuss here. The success of several recent models \cite{ouyangTrainingLanguageModels2022, openaiGPT4TechnicalReport, touvron2023llama} has prompted increased interest in reinforcement learning with human feedback \cite{christianoDeepReinforcementLearning2017, zieglerFineTuningLanguageModels2020}. While these models have improved performance on some pragmatic factors (i.e., following instructions) opportunities for pragmatic improvements remain. Such systems burden the user to specify what information should be considered. Furthermore, their performance notably differs from humans and the lack of transparency around these reasoning and language differences impairs their pragmatic %appropriateness
sufficiency (i.e., \cite{dasgupta_language_2022, sealsLongformAnalogiesGenerated2023}). 

We take an integrated approach designed to taxonomize recurrent themes, motivate a theoretical framework, and coordinate research efforts. We suggest that a unified framework facilitates integration with applied work on human expectations for conversational AI applications \cite{ashktorab_resilient_2019, liao_what_2016, luger_like_2016, zamora_im_2017}. Our framework integrates these issues with theoretical and empirical work in pragmatics.

\subsection{Limitations and Ethical Considerations}

This type of work inherits several limitations and ethical concerns related to the development of large models \cite{strubell_energy_2019} and privacy concerns common to conversational AI systems. Many external context features require information outside the lexical content of a conversation. %While many 
Some, but not all, users want systems to use this information \cite{luger_like_2016}, requiring customizable sharing settings. Moreover, we must avoid creating sub-optimal systems for users who share less information \cite{zuboff_age_2020}. Systems that request specific information may overcome this limitation. Second, our position could 
suggest an endorsement of larger models with high monetary and energy costs \cite{strubell_energy_2019}. However, pre-existing knowledge sources \cite{gaur__2018, miller_wordnet_1995, sheth_semantic_2005, speer_conceptnet_2017, valiant_knowledge_2006}, modular designs \cite{shuster_blenderbot_2022, leePromptedLLMsChatbot2023}, and approaches that address dialogue phenomena (i.e., \cite{benotti_grounding_2021, kim_deciding_2021, majumder_ask_2021}) are promising alternatives. Larger models alone will not resolve pragmatic limitations. While chatGPT improves on some tests posed by \cite{davis_experiments_2022}, clear limitations remain. %Truly pragmatically appropriate conversational AI systems will require coordinated approaches to addressing pragmatics.
Truly pragmatically-appropriate systems will require coordinated approaches that address multiple deficits. 

\section{Conclusion}

Several types of pragmatic challenges recur across current, disparate  conversational AI applications. We use examples from fielded conversational AI systems that are syntactically correct but have clear pragmatic deficiencies. These results contribute to a better understanding of the current pragmatic limitations of conversational AI systems. Moreover, they emphasize the importance of connections between general knowledge and the external environment in developing future conversational AI systems that better meet the pragmatic expectations of users.

\bibliographystyle{IEEEtran}
\bibliography{zotero}
\addtolength{\textheight}{-1cm}
\end{document}